\documentclass[11pt,a4paper]{article}
\usepackage[hyperref]{naaclhlt2018}
\usepackage{times}
\usepackage{latexsym}

\usepackage{url}

\usepackage{boxedminipage}
\usepackage{xcolor}

\usepackage{amsmath}
\usepackage{bm}

\DeclareMathOperator*{\softmax}{softmax}

\DeclareMathOperator{\dec}{dec}
\DeclareMathOperator{\enc}{enc}

\newcommand{\ascal}{\alpha}

\newcommand{\xvec}{\mathbf{x}}
\newcommand{\hvec}{\mathbf{h}}
\newcommand{\cvec}{\mathbf{c}}
\newcommand{\svec}{\mathbf{s}}
\newcommand{\yvec}{\mathbf{y}}

\newcommand{\xmat}{\mathbf{X}}
\newcommand{\hmat}{\mathbf{H}}
\newcommand{\cmat}{\mathbf{C}}
\newcommand{\smat}{\mathbf{S}}
\newcommand{\ymat}{\mathbf{Y}}
\newcommand{\amat}{\mathbf{A}}
\newcommand{\idmat}{\mathbf{I}}

\newcommand{\step}[1]{_{#1}}
\newcommand{\fst}{^1}
\newcommand{\snd}{^2}
\newcommand{\tri}{^{12}}

\newcommand{\uptimes}{\mathbin{\rotatebox[origin=c]{18}{$\rightarrow$}}}
\newcommand{\dtimes}{\mathbin{\rotatebox[origin=c]{-18}{$\rightarrow$}}}

\aclfinalcopy
\usepackage{multirow}
\usepackage{booktabs}

\usepackage{graphicx} %package to manage images
\graphicspath{ {figures/} }
\usepackage{subcaption}
\usepackage{rotating}

\DeclareCaptionFont{10pt}{\fontsize{10pt}{12pt}\selectfont}
\usepackage{tikz}
\usetikzlibrary{shapes.geometric}
\usepackage{pgfplots}

\usepackage{txfonts}

\title{Tied Multitask Learning for Neural Speech Translation}
\author{Antonios Anastasopoulos \and David Chiang \\
Department of Computer Science and Engineeering\\
University of Notre Dame \\
 {\tt \{aanastas,dchiang\}@nd.edu}
}
\date{}

\begin{document}

\maketitle

\begin{abstract}
     We explore multitask models for neural translation of speech, augmenting them in order to reflect two intuitive notions. First, we introduce a model where the second task decoder receives information from the decoder of the first task, since higher-level intermediate representations should provide useful information. Second, we apply regularization that encourages \textit{transitivity} and \textit{invertibility}. We show that the application of these notions on jointly trained models improves performance on the tasks of low-resource speech transcription and translation. It also leads to better performance when using attention information for word discovery over unsegmented input.
\end{abstract}

\section{Introduction}

Recent efforts in endangered language documentation focus on collecting spoken language resources, accompanied by spoken translations in a high resource language to make the resource interpretable \cite{bird-EtAl:2014:Coling}. For example, the BULB project \cite{adda2016breaking} used the LIG-Aikuma mobile app \cite{bird-EtAl:2014:W14-22,blachon2016parallel} to collect parallel speech corpora between three Bantu languages and French. 
Since it's common for speakers of endangered languages to speak one or more additional languages, collection of such a resource is a realistic goal.

Speech can be interpreted either by transcription in the original language or translation to another language. Since the size of the data is extremely small, multitask models that jointly train a model for both tasks can take advantage of both signals. Our contribution lies in improving the sequence-to-sequence multitask learning paradigm, by drawing on two intuitive notions: that higher-level representations are more useful than lower-level representations, and that translation should be both transitive and invertible.

\emph{Higher-level intermediate representations}, such as transcriptions, should in principle carry information useful for an end task like speech translation. A typical multitask setup \cite{weiss2017sequence} shares information at the level of encoded frames, but intuitively, a human translating speech must work from a higher level of representation, at least at the level of phonemes if not syntax or semantics.
Thus, we present a novel architecture for \emph{tied} multitask learning with sequence-to-sequence models, in which the decoder of the second task receives information not only from the encoder, but also from the decoder of the first task.

In addition, \textit{transitivity} and \textit{invertibility} are two properties that should hold when mapping between levels of representation or across languages. We demonstrate how these two notions can be implemented through regularization of the attention matrices, and how they lead to further improved performance.

We evaluate our models in three experiment settings: low-resource speech transcription and translation, word discovery on unsegmented input, and high-resource text translation.
Our high-resource experiments are performed on English, French, and German. Our low-resource speech experiments cover a wider range of linguistic diversity: Spanish-English, Mboshi-French, and Ainu-English.

In the speech transcription and translation tasks, our proposed model leads to improved performance against all baselines as well as previous multitask architectures. We observe improvements of up to $5\%$ character error rate in the transcription task, and up to $2.8\%$ character-level BLEU in the translation task. 
However, we didn't observe similar improvements in the text translation experiments. 
Finally, on the word discovery task, we
improve upon previous work by about $3\%$ F-score on both tokens and types.

\section{Model}

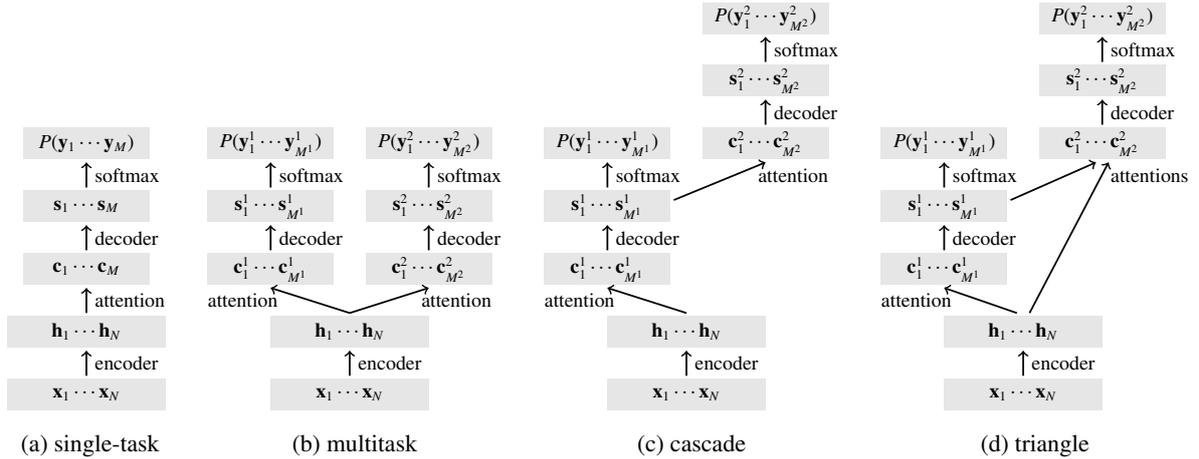
\begin{figure*}
\tikzset{seq/.style={draw=none,fill=gray!20}}
\tikzset{layer/.style={->,thick}}
\tikzset{label/.style={anchor=west,font={\footnotesize}}}
\tikzset{seqlabel/.style={font={\small}}}
\newcommand{\encoder}{
\draw[seq] (-1.25,-0.25) rectangle (1.25,0.25);
\node[seqlabel] at (0,0) 
%{$\xmat$};
{$\xvec\step1 \cdots \xvec\step{N}$};
\draw[layer] (0,0.3) -- (0,0.7);
\node[seqlabel] at (0,0.5) [label] {encoder};
\draw[seq] (-1.25,0.75) rectangle (1.25,1.25);
\node[seqlabel] at (0,1) %{$\hmat$}; 
{$\hvec\step1 \cdots \hvec\step{N}$};
}
\newcommand{\decoder}[1]{
\draw[seq] (-1,1.75) rectangle (1,2.25);
\node[seqlabel] at (0,2) %{$\cmat#1$}; 
{$\cvec\step{1}#1 \cdots \cvec\step{M#1}#1$};
\draw[layer] (0,2.3) -- (0,2.7);
\node[seqlabel] at (0,2.5) [label] {decoder};
\draw[seq] (-1,2.75) rectangle (1,3.25);
\node[seqlabel] at (0,3) %{$\smat#1$}; 
{$\svec\step{1}#1 \cdots \svec\step{M#1}#1$};
\draw[layer] (0,3.3) -- (0,3.7);
\node[seqlabel] at (0,3.5) [label] {softmax};
\draw[seq] (-1,3.75) rectangle (1,4.25);
\node[seqlabel] at (0,4) %{$P(\ymat#1)$}; 
{$P(\yvec\step{1}#1 \cdots \yvec\step{M#1}#1)$};
}
\begin{center}
\resizebox{\hsize}{!}{%
\begin{tabular}{cccc}
\begin{tikzpicture}
\encoder
\draw[layer] (0,1.3) -- (0,1.7);
\node at (0,1.5) [label] {attention};
\decoder{}
\end{tikzpicture}
&
\begin{tikzpicture}
\encoder
\draw[layer] (0,1.3) -- (-1.25,1.7);
\draw[layer] (0,1.3) -- (1.25,1.7);
\node at (-1,1.5) [label,anchor=east] {attention};
\node at (1,1.5) [label] {attention};
\begin{scope}[xshift=-1.25cm]
\decoder{^1}
\end{scope}
\begin{scope}[xshift=1.25cm]
\decoder{^2}
\end{scope}
\end{tikzpicture}
&
\begin{tikzpicture}
\encoder
\draw[layer] (0,1.3) -- (-1.25,1.7);
\node at (-1,1.5) [label,anchor=east] {attention};
\begin{scope}[xshift=-1.25cm]
\decoder{^1}
\end{scope}
\draw[layer] (-0.2,3.1) -- (1.25,3.7);
\node at (1,3.5) [label] {attention};
\begin{scope}[xshift=1.25cm,yshift=2cm]
\decoder{^2}
\end{scope}
\end{tikzpicture}
&
\begin{tikzpicture}
\encoder
\draw[layer] (-0.1,1.3) -- (-1.25,1.7);
\node at (-1,1.5) [label,anchor=east] {attention};
\begin{scope}[xshift=-1.25cm]
\decoder{^1}
\end{scope}
\draw[layer] (-0.2,3.1) -- (1.15,3.7);
\node at (1.25,3.5) [label] {attentions};
\draw[layer] (0.1,1.3) -- (1.35,3.7);
\begin{scope}[xshift=1.25cm,yshift=2cm]
\decoder{^2}
\end{scope}
\end{tikzpicture}
\\[1ex]
(a) single-task & (b) multitask & (c) cascade & (d) triangle
\end{tabular}%
}
\end{center}
\caption{Variations on the standard attentional model. In the standard \emph{single-task} model, the decoder attends to the encoder's states. In a typical \emph{multitask} setup, two decoders attend to the encoder's states. In the \emph{cascade} \cite{tu2017neural}, the second decoder attends to the first decoder's states. In our proposed \emph{triangle} model, the second decoder attends to both the encoder's states and the first decoder's states. Note that for clarity's sake there are dependencies not shown.}
\label{fig:models}
\end{figure*}

Our models are based on a sequence-to-sequence model with attention \cite{bahdanau2014}. In general, this type of model is composed of three parts: a recurrent encoder, the attention, and a recurrent decoder (see Figure~\ref{fig:models}a).\footnote{For simplicity, we have assumed only a single layer for both the encoder and decoder. It is possible to use multiple stacked RNNs; typically, the output of the encoder and decoder ($\cvec\step{m}$ and $P(\yvec\step{m})$, respectively) would be computed from the top layer only.}

The encoder transforms an input sequence of words or feature frames $\xvec\step{1}, \ldots, \xvec\step{N}$ into a sequence of \emph{input states} $\hvec\step{1}, \ldots, \hvec\step{N}$:
\[
\hvec\step{n} = \enc(\hvec\step{n-1}, \xvec\step{n}).
\]
The attention transforms the input states into a sequence of \emph{context vectors} via a matrix of \emph{attention weights}:
\[
\cvec\step{m} = \sum_n \ascal_{mn} \hvec\step{n}.
\]
Finally, the decoder computes a sequence of \emph{output states} from which a probability distribution over output words can be computed.
\begin{align*}
\svec\step{m} &= \dec(\svec\step{m-1}, \cvec\step{m}, \yvec\step{m-1}) \\
P(\yvec\step{m}) &= \softmax(\svec\step{m}).
\end{align*}

In a standard encoder-decoder \emph{multitask} model (Figure~\ref{fig:models}b) \cite{dongetal2015multitask, weiss2017sequence}, we jointly model two output sequences using a shared encoder, but separate attentions and decoders:
\begin{align*}
\cvec\step{m}\fst &= \sum_n \ascal\fst_{mn} \hvec\step{n} \\ \svec\step{m}\fst &= \dec\fst(\svec\step{m-1}\fst, \cvec\step{m}\fst, \yvec\step{m-1}\fst) \\
P(\yvec\step{m}\fst) &= \softmax(\svec\step{m}\fst)  \\
\intertext{and}
\cvec\step{m}\snd &= \sum_n \ascal\snd_{mn} \hvec\step{n} \\
\svec\step{m}\snd &= \dec\snd(\svec\step{m-1}\snd, \cvec\step{m}\snd, \yvec\step{m-1}\snd) \\
P(\yvec\step{m}\snd) &= \softmax(\svec\step{m}\snd).
\end{align*}

We can also arrange the decoders in a \emph{cascade} (Figure~\ref{fig:models}c), in which the second decoder attends only to the output states of the first decoder: 
\begin{align*}
\cvec\step{m}\snd &= \sum_{m'} \ascal_{mm'}\tri \svec\step{m'}\fst \\
\svec\step{m}\snd &= \dec\snd(\svec\step{m-1}\snd, \cvec\step{m}\snd, \yvec\step{m-1}\snd) \\
P(\yvec\step{m}\snd) &= \softmax(\svec\step{m}\snd).
\end{align*}
\citet{tu2017neural} use exactly this architecture to train on bitext by setting the second output sequence to be equal to the input sequence ($\yvec\step{i}\snd = \xvec\step{i}$).

In our proposed \emph{triangle} model (Figure~\ref{fig:models}d), the first decoder is as above, but the second decoder has two attentions, one for the input states of the encoder and one for the output states of the first decoder:
\begin{align*}
\cvec\step{m}\snd &= \begin{bmatrix} \sum_{m'} \ascal_{mm'}\tri \svec\step{m'}\fst & \sum_n \ascal_{mn}\snd \hvec\step{n} \end{bmatrix} \\
\svec\step{m}\snd &= \dec\snd(\svec\step{m-1}\snd, \cvec\step{m}\snd, \yvec\step{m-1}\snd) \\
P(\yvec\step{m}\snd) &= \softmax(\svec\step{m}\snd).
\end{align*}
Note that the context vectors resulting from the two attentions are concatenated, not added.

\section{Learning and Inference}

For compactness, we will write $\xmat$ for the matrix whose rows are the $\xvec\step{n}$, and similarly $\hmat$, $\cmat$, and so on. We also write $\amat$ for the matrix of attention weights: $[\amat]_{ij} = \ascal_{ij}$.

Let $\theta$ be the parameters of our model, which we train on sentence triples $(\xmat, \ymat\fst, \ymat\snd)$.

\subsection{Maximum likelihood estimation}

Define the score of a sentence triple to be a log-linear interpolation of the two decoders' probabilities:
\begin{align*}
\text{score}(\ymat\fst, \ymat\snd \mid \xmat; \theta) &= \lambda\log P(\ymat\fst \mid \xmat ; \theta) + {} \\
& \quad (1-\lambda) \log P(\ymat\snd \mid \xmat, \smat\fst ; \theta) 
\end{align*}
where $\lambda$ is a parameter that controls the importance of each sub-task. In all our experiments, we set $\lambda$ to $0.5$.
We then train the model to maximize
\begin{equation*}
\mathcal{L}(\theta) = \sum \text{score}(\ymat\fst, \ymat\snd \mid  \xmat; \theta),
\end{equation*}
where the summation is over all sentence triples in the training data.

\subsection{Regularization}

We can optionally add a regularization term to the objective function, in order to encourage our attention mechanisms to conform to two intuitive principles of machine translation: \textit{transitivity} and \textit{invertibility}.

\paragraph{\emph{Transitivity} attention regularizer}

To a first approximation, the translation relation should be transitive \cite{wang+al:2006,levinboim2015multi}: If source word $\xvec\step{i}$ aligns to target word $\yvec\fst\step{j}$ and $\yvec\fst\step{j}$ aligns to target word $\yvec\snd\step{k}$, then $\xvec\step{i}$ should also probably align to $\yvec\snd\step{k}$. To encourage the model to preserve this relationship, we add the following \textit{transitivity} regularizer to the loss function of the \emph{triangle} models with a small weight $\lambda_{\text{trans}} = 0.2$: 
\[
\mathcal{L}_{\text{trans}} = \text{score}(\ymat\fst, \ymat\snd) - \lambda_{\text{trans}} \, \bigl\|  \amat\tri\amat\fst - \amat\snd \bigr\|_2^2 .
\]

\paragraph{\textit{Invertibility} attention regularizer}

The translation relation also ought to be roughly invertible  \cite{levinboim2015model}: if, in the \emph{reconstruction} version of the \emph{cascade} model, source word $\xvec\step{i}$ aligns to target word $\yvec\step{j}\fst$, then it stands to reason that $\yvec\step{j}$ is likely to align to $\xvec\step{i}$. So,  whereas \citet{tu2017neural} let the attentions of the translator and the reconstructor be unrelated, we try adding the following \textit{invertibility} regularizer to encourage the attentions to each be the inverse of the other, again with a weight $\lambda_{\text{inv}}=0.2$:
\[
\mathcal{L}_{\text{inv}} = \text{score}(\ymat\fst, \ymat\snd) - \lambda_{\text{inv}} \, \bigl\| \amat\fst\amat\tri - \idmat \bigr\|_2^2 .
\]

\subsection{Decoding}

Since we have two decoders, we now need to employ a two-phase beam search, following \citet{tu2017neural}:
\begin{enumerate}
    \item The \textit{first decoder} produces, through standard beam search, a set of triples each consisting of a candidate transcription $\hat{\ymat}\fst$, a score $P(\hat{\ymat}\fst)$, and a hidden state sequence $\hat{\smat}$.
    \item For each transcription candidate from the \textit{first decoder}, the \textit{second decoder} now produces through beam search a set of candidate translations $\hat{\ymat}\snd$, each with a score $P(\hat{\ymat}\snd)$.
    \item We then output the combination that yields the highest total $\text{score}(\ymat\fst, \ymat\snd)$.
\end{enumerate}

\subsection{Implementation}

All our models are implemented in DyNet \cite{neubig2017dynet}.\footnote{Our code is available at: \url{https://bitbucket.org/antonis/dynet-multitask-models}.} We use a dropout of 0.2, and train using Adam with initial learning rate of $0.0002$ for a maximum of 500 epochs. For testing, we select the model with the best performance on dev. At inference time, we use a beam size of 4 for each decoder (due to GPU memory constraints), and the beam scores include length normalization \cite{wu2016google} with a weight of 0.8, which \citet{nguyen2017transfer} found to work well for low-resource NMT.

\section{Speech Transcription and Translation}
\label{sec:st}

\begin{table}[]
    \centering
    \begin{tabular}{@{}l|ccc@{}}
    \toprule
        Corpus & Speakers & Segments & Hours \\
    \midrule
        Ainu-English & 1 & 2,668 & 2.5\\
        Mboshi-French & 3 & 5,131 & 4.4\\
        Spanish-English & 240 & 17,394 & 20 \\
    \bottomrule
    \end{tabular}
    \caption{Statistics on our speech datasets.}
    \label{tab:speechsorpora}
\end{table}
\begin{table*}[h]
    \centering
    \begin{tabular}{c|cc|cc|cc|cc|cc}
    \toprule
    &\multicolumn{2}{c|}{Model} & \multicolumn{2}{c|}{Search}  & Mboshi & French & Ainu & English & Spanish & English \\
    &ASR & MT & ASR & MT & CER & BLEU & CER & BLEU & CER & BLEU\\
    \midrule
        (1) & auto & text & 1-best & 1-best & 42.3 & 21.4 & 44.0 & 16.4 & 63.2 & 24.2 \\
        (2) & gold & text & --- & 1-best & ~~0.0 & 31.2 & ~~0.0 & 19.3 & ~~0.0 & 51.3 \\
        (3) & \multicolumn{2}{c|}{single-task} & \multicolumn{2}{c|}{1-best} & --- & 20.8 & --- & 12.0 & --- & 21.6 \\
        (4) & \multicolumn{2}{c|}{multitask} & 4-best & 1-best & 36.9 & 21.0 & 40.1 & 18.3 & \textbf{57.4} & 26.0  \\
    \midrule
        (5) & \multicolumn{2}{c|}{cascade} & 4-best & 1-best & 39.7 & 24.3  & 42.1 & 19.8 & 58.1 & 26.8 \\
    \midrule
        (6) & \multicolumn{2}{c|}{triangle} & 4-best & 1-best & 32.3 & 24.1  & 39.9 & 19.2 & 58.9 & \textbf{28.6} \\

        (7) & \multicolumn{2}{c|}{triangle+$\mathcal{L}_{\text{trans}}$} & 4-best & 1-best & 33.0 & \textbf{24.7} & 43.3 & \textbf{20.2} & 59.3 & \textbf{28.6} \\
    \midrule
        (8) & \multicolumn{2}{c|}{triangle} & 1-best & 1-best & \textbf{31.8} & 19.7 & \textbf{38.9} & \textbf{19.8} & 58.4 & \textbf{28.8} \\
        (9) & \multicolumn{2}{c|}{triangle+$\mathcal{L}_{\text{trans}}$} & 1-best & 1-best & 32.1 & 20.9 & 43.0 &\textbf{20.3} & 59.1 & \textbf{28.5} \\
    \bottomrule
    \end{tabular}
    \caption{The multitask models outperform the baseline single-task model and the pivot approach (auto/text) on all language pairs tested. The \emph{triangle} model also outperforms the simple multitask models on both tasks in almost all cases. The best results for each dataset and task are highlighted.}
    \label{tab:speech}
\end{table*}

We focus on speech transcription and translation of endangered languages, using three different corpora on three different language directions: Spanish (es) to English (en), Ainu (ai) to English, and Mboshi (mb) to French (fr).

\subsection{Data}

Spanish is, of course, not an endangered language, but the availability of the CALLHOME Spanish Speech dataset (LDC2014T23) with English translations \cite{post2013improved} makes it a convenient language to work with, as has been done in almost all previous work in this area. It consists of telephone conversations between relatives (about~20 total hours of audio) with more than 240 speakers. We use the original train-dev-test split, with the training set comprised of 80 conversations and dev and test of 20 conversations each.

Hokkaido Ainu is the sole surviving member of the Ainu language family and is generally considered a language isolate. As of 2007, only ten native speakers were alive. The Glossed Audio Corpus of Ainu Folklore 
provides 10 narratives with audio (about~2.5 hours of audio) and translations in Japanese and English.\footnote{\url{http://ainucorpus.ninjal.ac.jp/corpus/en/}} Since there does not exist a standard train-dev-test split, we employ a cross validation scheme for evaluation purposes. In each fold, one of the 10 narratives becomes the test set, with the previous one (mod 10) becoming the dev set, and the remaining 8 narratives becoming the training set. The models for each of the~10 folds are trained and tested separately. On average, for each fold, we train on about~2000 utterances; the dev and test sets consist of about~$270$ utterances. We report results on the concatenation of all folds.
The Ainu text is split into characters, except for the equals (\texttt=) and underscore (\texttt\_) characters, which are used as phonological or structural markers and are thus merged with the following character.\footnote{The data preprocessing scripts are released with the rest of our code.}

Mboshi (Bantu C25 in the Guthrie classification) is a language spoken in Congo-Brazzaville, without standard orthography. We use a corpus \cite{godard2017very} of 5517 parallel utterances (about~4.4 hours of audio) collected from three native speakers. The corpus provides non-standard grapheme transcriptions (close to the language phonology) produced by linguists, as well as French translations. 
We sampled~100 segments from the training set to be our dev set, and used the original dev set (514 sentences) as our test set. 

\subsection{Implementation}

We employ a 3-layer speech encoding scheme similar to that of \citet{long-EtAl:2016:NAACL-HLT}. The first bidirectional layer receives the audio sequence in the form of 39-dimensional Perceptual Linear Predictive (PLP) features \cite{hermansky1990perceptual} computed over overlapping 25ms-wide windows every 10ms. The second and third layers consist of LSTMs with hidden state sizes of 128 and 512 respectively. Each layer encodes every second output of the previous layer.
Thus, the sequence is downsampled by a factor of 4, decreasing the computation load for the attention mechanism and the decoders. In the speech experiments, the decoders output the sequences at the grapheme level, so the output embedding size is set to 64.

We found that this simpler speech encoder works well for our extremely small datasets. Applying our models to larger datasets with many more speakers would most likely require a more sophisticated speech encoder, such as the one used by \citet{weiss2017sequence}.

\subsection{Results}

\begin{figure*}[h]
\resizebox{\hsize}{!}{%
\!\!\!\!
\begin{tabular}{@{}ccc@{}}
$\amat^1$ & $\amat^1$ &\\ \includegraphics[width=2.5in]{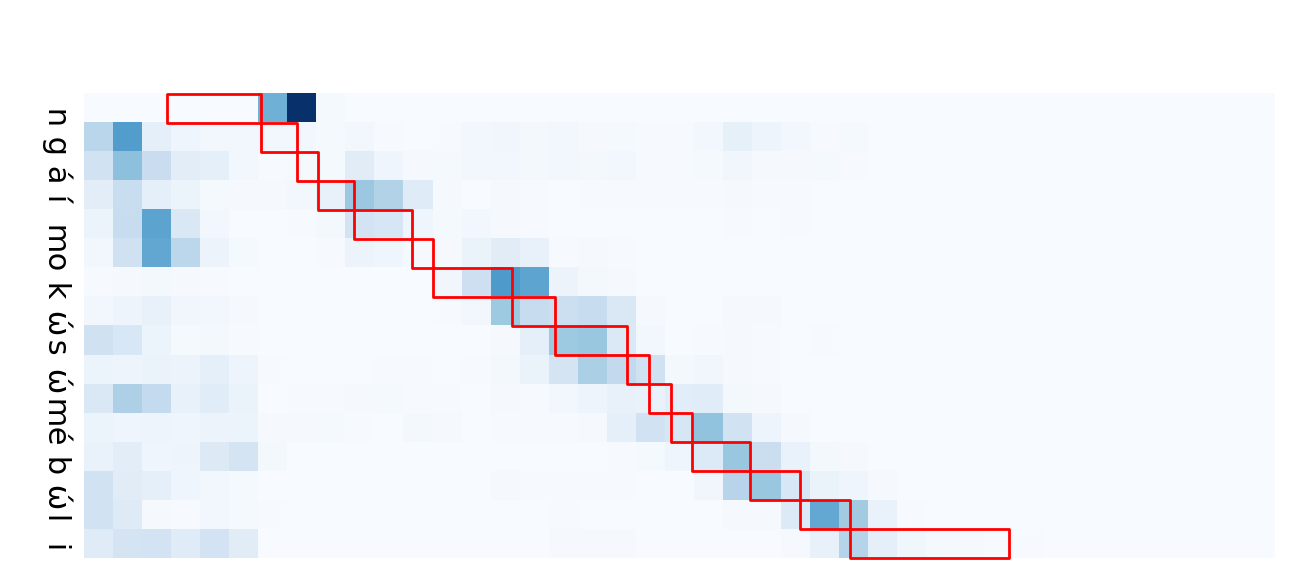}
&
\includegraphics[width=2.5in]{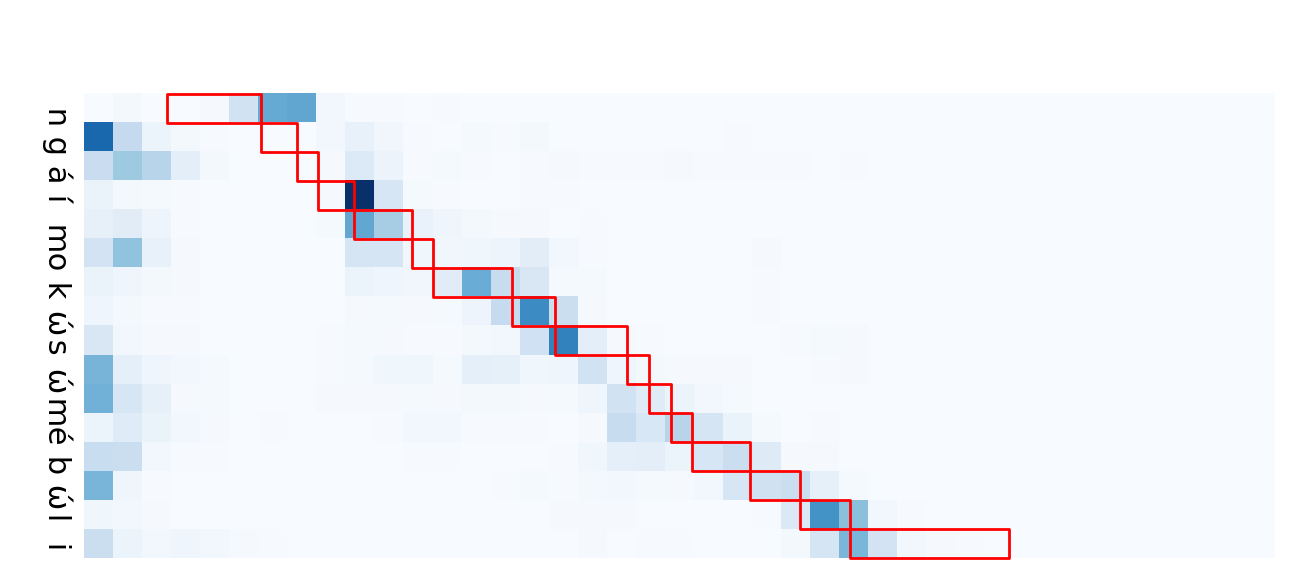} &\\

\smash{\raisebox{-.55\normalbaselineskip}{$\amat^2$}} & \smash{\raisebox{-.55\normalbaselineskip}{$\amat^2$}} & \smash{\raisebox{-.1\normalbaselineskip}{$\amat^{12}$}}\\

\includegraphics[width=2.5in]{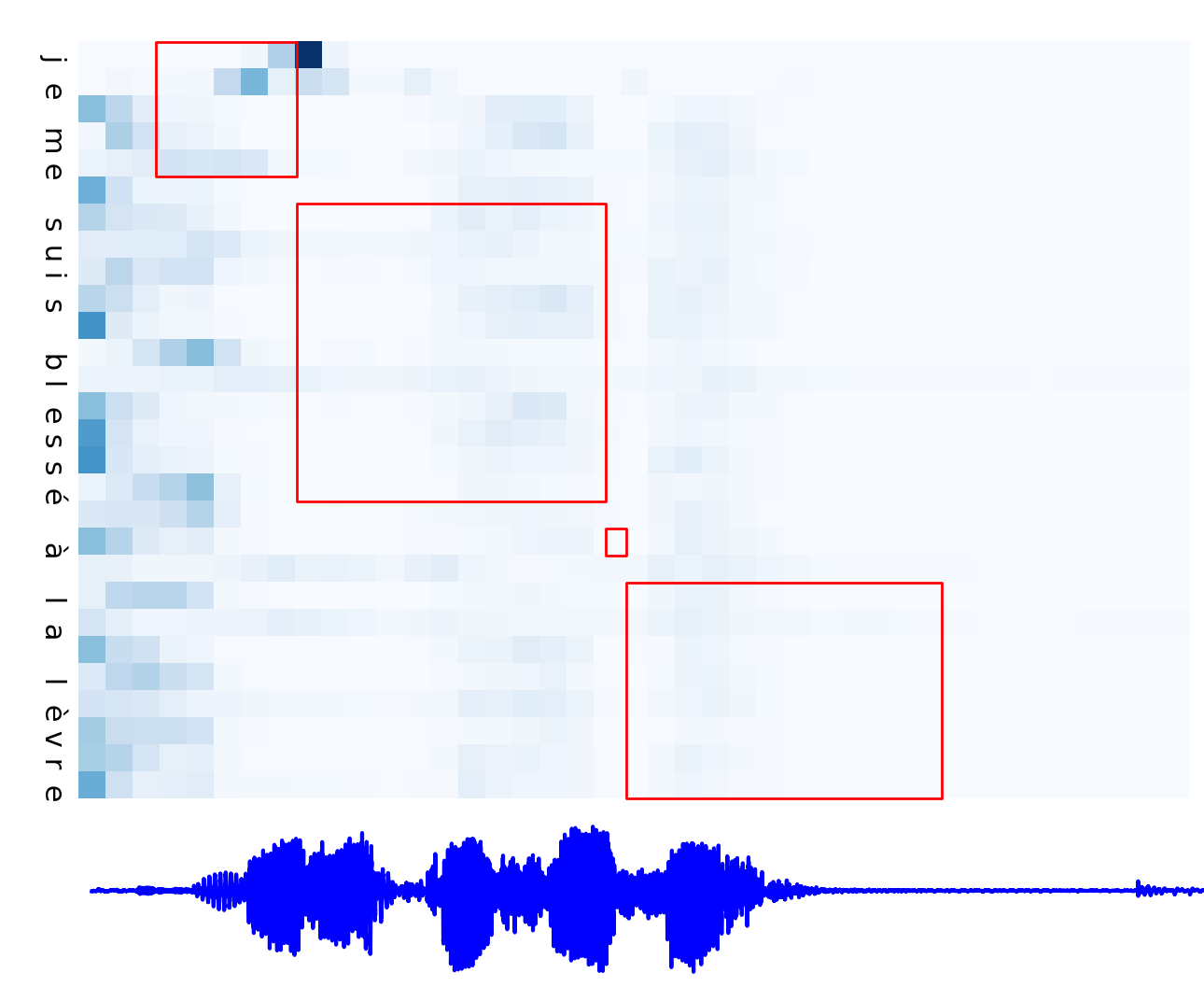}
&
\includegraphics[width=2.5in]{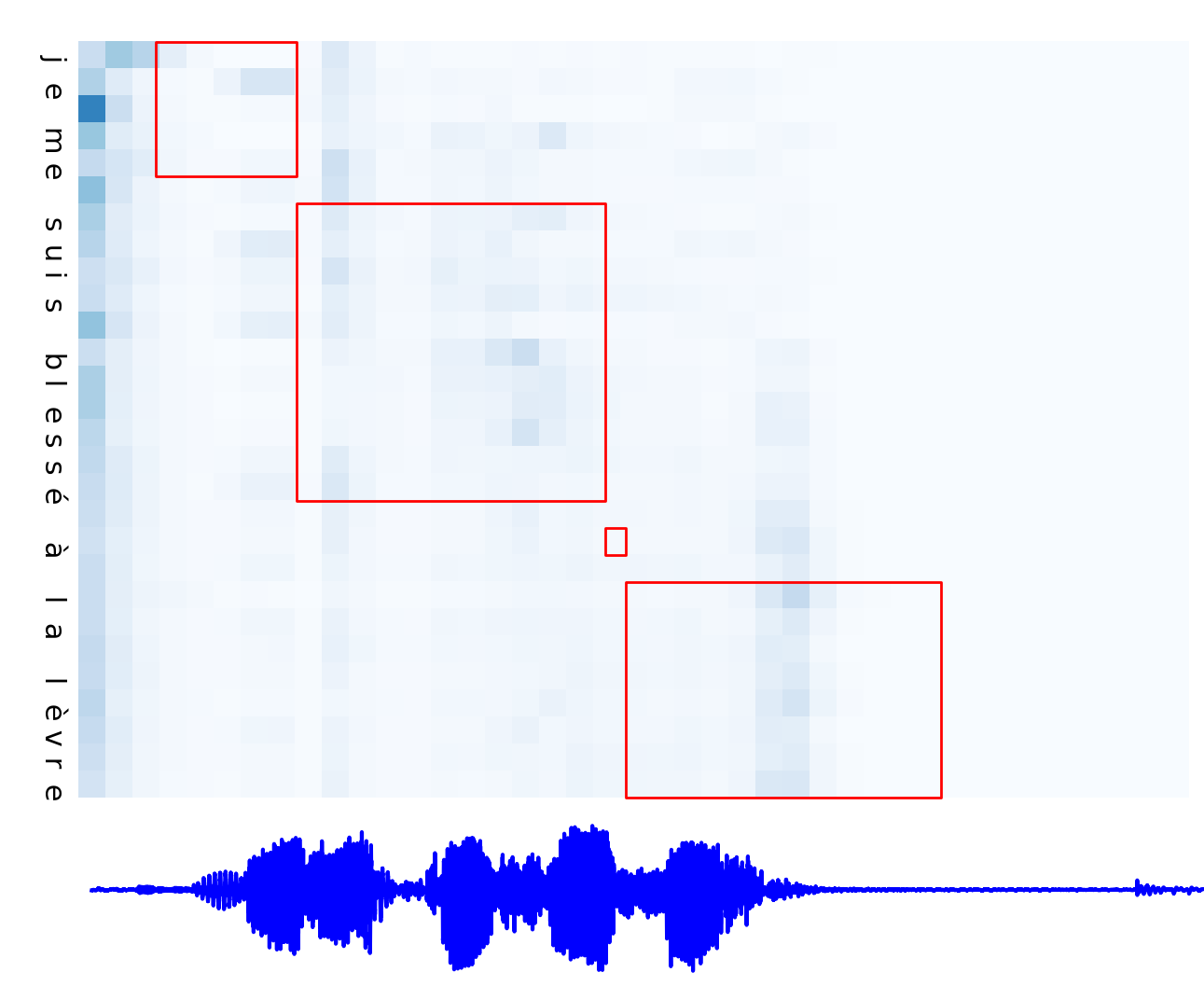}
&
\raisebox{2\normalbaselineskip}{\includegraphics[width=1in]{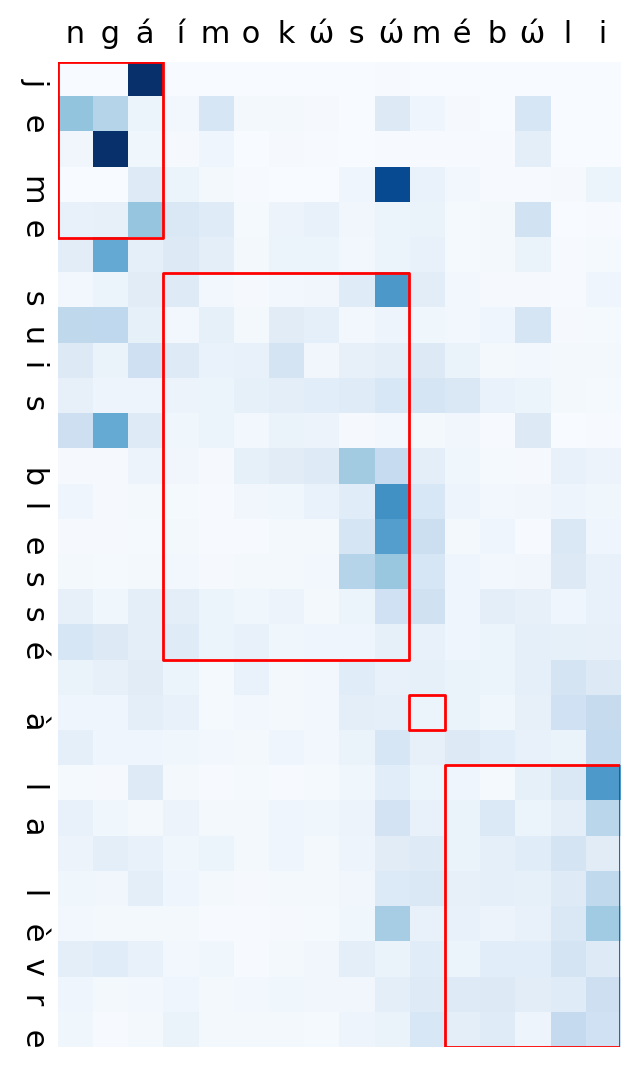}}
\\

(a) multitask & \multicolumn{2}{c}{(b) triangle + transitivity}\\
\end{tabular}
}
\caption{Attentions in an Mboshi-French sentence, extracted from two of our models. The red squares denote gold alignments. The second decoder of the \emph{triangle} model receives most of its context from the first decoder through $\amat^{12}$ instead of the source. The $\amat^2$ matrix of the \textit{triangle} model is more informed (34\% correct attention mass) than the \textit{multitask} one (21\% correct), due to the \textit{transitivity} regularizer.} 
\label{fig:attentions}
\end{figure*}

In Table~\ref{tab:speech}, we present results on three small datasets that demonstrate the efficacy of our models. We compare our proposed models against three baselines and one ``skyline.'' The first baseline is a traditional pivot approach (line 1), where the ASR output, a sequence of characters, is the input to a character-based NMT system (trained on gold transcriptions). The ``skyline'' model (line 2) is the same NMT system, but tested on gold transcriptions instead of ASR output. The second baseline is translation directly from source speech to target text (line 3). The last baseline is the standard \emph{multitask} model (line 4), which is similar to the model of \citet{weiss2017sequence}.

The \emph{cascade} model (line 5) outperforms the baselines on the translation task, while only falling behind the \emph{multitask} model in the transcription task.
On all three datasets, the \emph{triangle} model (lines 6, 7) outperforms all baselines, including the standard \emph{multitask} model. On Ainu-English, we even obtain translations that are comparable to the ``skyline'' model, which is tested on gold Ainu transcriptions.

Comparing the performance of all models across the three datasets, there are two notable trends that verify common intuitions regarding the speech transcription and translation tasks. First, an increase in the number of speakers hurts the performance of the speech transcription tasks. The character error rates for Ainu are smaller than the CER in Mboshi, which in turn are smaller than the CER in CALLHOME. Second, the character-level BLEU scores increase as the amount of training data increases, with our smallest dataset (Ainu) having the lowest BLEU scores, and the largest dataset (CALLHOME) having the highest BLEU scores. This is expected, as more training data means that the translation decoder learns a more informed character-level language model for the target language. 

Note that \citet{weiss2017sequence} report much higher BLEU scores on CALLHOME: our model underperforms theirs by almost~9 \textit{word-level} BLEU points. However, their model has significantly more parameters and is trained on~10 times more data than ours. Such an amount of data would never be available in our endangered languages scenario. When calculated on the word-level, all our models' BLEU scores are between~3 and~7 points for the extremely low resource datasets (Mboshi-French and Ainu-English), and between~7 and~10 for CALLHOME. Clearly, the size of the training data in our experiments is not enough for producing high quality speech translations, but we plan to investigate the performance of our proposed models on larger datasets as part of our future work.

To evaluate the effect of using the combined score from both decoders at decoding time, we evaluated the \emph{triangle} models using only the \mbox{1-best} output from the speech model (lines 8, 9). One would expect that this would favor speech at the expense of translation. In transcription accuracy, we indeed observed improvements across the board. In translation accuracy, we observed a surprisingly large drop on Mboshi-French, but surprisingly little effect on the other language pairs -- in fact, BLEU scores tended to go up slightly, but not significantly.

Finally, Figure~\ref{fig:attentions} visualizes the attention matrices for one utterance from the baseline multitask model and our proposed \emph{triangle} model. It is clear that our intuition was correct: the translation decoder receives most of its context from the transcription decoder, as indicated by the higher attention weights of $\amat^{12}$. Ideally, the area under the red squares (gold alignments) would account for~100\% of the attention mass of $\amat^{12}$. In our triangle model, the total mass under the red squares is~34\%, whereas the multitask model's correct attentions amount to only~21\% of the attention mass.

\section{Word Discovery}
\label{sec:worddisc}
\begin{table*}[]
    \centering
    \begin{tabular}{lc|rrr|rrr}
    \toprule
    \multicolumn{2}{l}{\multirow{2}{*}{Model (with smoothing)}}   & \multicolumn{3}{|c}{Tokens} & \multicolumn{3}{|c}{Types}  \\
        &    & Precision & Recall & F-score & Precision & Recall & F-score \\
    \midrule 
         \citealt{boito2017unwritten} & \textit{base} & \textit{5.85} & \textit{6.82} & \textit{6.30} & \textit{6.76} & \textit{15.00} & \textit{9.32} \\
         (reported) & \textit{reverse} & \textit{21.44} & \textit{16.49} & \textit{18.64} & \textit{27.23} & \textit{15.02} & \textit{19.36} \\
    \midrule
        \citealt{boito2017unwritten} & base & 6.87 & 6.33 & 6.59 & 6.17 & 13.02 & 8.37 \\
        (reimplementation) & reverse & 7.58 & 8.16 & 7.86 & 9.22 & 11.97 & 10.42\\
    \midrule
        \multirow{2}{*}{our single-task} & base & 7.99 & 7.57 & 7.78 & 7.59 & \textbf{16.41} & 10.38\\
        & reverse & \textbf{11.31} & \textbf{11.82} & \textbf{11.56} & 9.29 & 14.75 & 11.40\\
    \midrule
        \multicolumn{2}{l|}{reconstruction${}+0.2\mathcal{L}_{\text{inv}}$}  &  8.93 & 9.78 & 9.33 & 8.66 & 15.48 & 11.02 \\
        \multicolumn{2}{l|}{reconstruction${}+0.5\mathcal{L}_{\text{inv}}$}  &   7.42 & 10.00 & 8.52 & \textbf{10.46} & 16.36 & \textbf{12.76} \\
    \bottomrule
    \end{tabular}
    \caption{The reconstruction model with the \emph{invertibility} regularizer produces more informed attentions that result in better word discovery for Mboshi with an Mboshi-French model. Scores reported by previous work are in \textit{italics} and best scores from our experiments are in \textbf{bold}.}
    \label{tab:worddisc}
\end{table*}

Although the above results show that our model gives large performance improvements, in absolute terms, its performance on such low-resource tasks leaves a lot of room for future improvement.
A possible more realistic application of our methods is word discovery, that is, finding word boundaries in unsegmented phonetic transcriptions.

After training an attentional encoder-decoder model between Mboshi unsegmented phonetic sequences and French word sequences, the attention weights can be thought of as soft alignments, which allow us to project the French word boundaries onto Mboshi. Although we could in principle perform word discovery directly on speech, we leave this for future work, and only explore single-task and reconstruction models.

\subsection{Data}
We use the same Mboshi-French corpus as in Section~\ref{sec:st}, but with the original training set of 4617 utterances and the dev set of 514 utterances. Our parallel data consist of the unsegmented phonetic Mboshi transcriptions, along with the word-level French translations. 

\subsection{Implementation}

We first replicate the model of \citet{boito2017unwritten}, with a single-layer bidirectional encoder and single layer decoder, using an embedding and hidden size of 12 for the base model, and an embedding and hidden state size of 64 for the reverse model. In our own models, we set the embedding size to 32 for Mboshi characters, 64 for French words, and the hidden state size to 64.
We smooth the attention weights $\mathbf{A}$ using the method of \citet{long-EtAl:2016:NAACL-HLT} with a temperature $T=10$ for the softmax computation of the attention mechanism.
%\[
%\alpha'_{mn} = \frac{\exp (\alpha_{mn}/T)}{\sum_k \exp ( \alpha_{mk}/T) }
%\]

Following \citet{boito2017unwritten}, we train models both on the \textit{base} Mboshi-to-French direction, as well as the \textit{reverse} (French-to-Mboshi) direction, with and without this smoothing operation.
We further smooth the computed soft alignments of all models so that $a_{mn} = (a_{mn-1} + a_{mn} + a_{mn+1})/3$ as a post-processing step. From the \textit{single-task} models we extract the $\mathbf{A}^1$ attention matrices. We also train \emph{reconstruction} models on both directions, with and without the \emph{invertibility} regularizer, extracting both $\mathbf{A}^1$ and $\mathbf{A}^{12}$ matrices. The two matrices are then combined so that $\amat = \amat^1 + (\amat^{12})^T$.

\subsection{Results}
Evaluation is done both at the token and the type level, by computing precision, recall, and F-score over the discovered segmentation, with the best results shown in Table~\ref{tab:worddisc}.
We reimplemented the base (Mboshi-French) and reverse (French-Mboshi) models from \citet{boito2017unwritten}, and the performance of the base model was comparable to the one reported. However, we were unable to reproduce the significant gains that were reported when using the reverse model (\textit{italicized} in Table~\ref{tab:worddisc}). Also, our version of both the base and reverse singletask models performed better than our reimplementation of the baseline. 

Furthermore, we found that we were able to obtain even better performance at the type level by combining the attention matrices of a reconstruction model trained with the \textit{invertibility} regularizer. \citet{boito2017unwritten} reported that combining the attention matrices of a base and a reverse model significantly reduced performance, but they trained the two models separately. In contrast, we obtain the base ($\amat^1$) and the reverse attention matrices ($\amat^{12}$) from a model that trains them jointly, while also tying them together through the \textit{invertibility} regularizer. Using the regularizer is key to the improvements; in fact, we did not observe any improvements when we trained the reconstruction models without the regularizer.

\section{Negative Results: High-Resource Text Translation}

\subsection{Data}
For evaluating our models on text translation, we use the Europarl corpus which provides parallel sentences across several European languages. We extracted 1,450,890 three-way parallel sentences on English, French, and German. The concatenation of the newstest 2011--2013 sets (8,017 sentences) is our dev set, and our test set is the concatenation of the newstest 2014 and 2015 sets (6,003 sentences). We test all architectures on the six possible translation directions between English (en), French (fr) and German (de). All the sequences are represented by subword units with byte-pair encoding (BPE) \cite{sennrich2016} trained on each language with $32000$ operations. 

\subsection{Experimental Setup}
On all experiments, the encoder and the decoder(s) have 2 layers of LSTM units with hidden state size and attention size of 1024, and embedding size of 1024. For this high resource scenario, we only train for a maximum of~40 epochs.

\subsection{Results}

The accuracy of all the models on all six language pair directions is shown in Table~\ref{tab:bigtable}.
\begin{table*}[t]
    \centering
    \begin{tabular}{l|cccccc}
    \toprule
    
        \multirow{2}{*}{Model} & \multicolumn{6}{|c}{$s\rightarrow t$} \\
         & en$\rightarrow$fr & en$\rightarrow$de & fr$\rightarrow$en & fr$\rightarrow$de & de$\rightarrow$en & de$\rightarrow$fr \\
    \midrule
        singletask  & \textbf{20.92} & \textbf{12.69} & \textbf{20.96} & \textbf{11.24} & \textbf{16.10} & \textbf{15.29} \\
        multitask $s \rightarrow x,t$  & 20.54 & \textbf{12.79} & 20.01 & \textbf{11.18} & \textbf{16.31} & \textbf{15.07} \\
        cascade $s\rightarrow x \rightarrow t$ & 15.93 & 11.31 & 16.58 & 7.60 & 13.46 & 13.24 \\
        cascade $s\rightarrow t \rightarrow x$ & 20.34 & 12.27 & 19.17 & \textbf{11.09} & 15.24 & 14.78 \\
        reconstruction & 20.19 & \textbf{12.44} & 20.63 & 10.88 & 15.66 & 13.44 \\
        reconstruction $+\mathcal{L}_{\text{inv}}$ & \textbf{20.72} & \textbf{12.64} & 20.11 & 10.46 & 15.43 & 12.64 \\
        triangle $s$ \raisebox{-2pt}{$\xrightarrow{\uptimes \scalebox{1}{\textit{x}} \dtimes}$} $t$ & 20.39 & \textbf{12.70} & 17.93 & 10.17 & 14.94 & 14.07 \\
        triangle $s$ \raisebox{-2pt}{$\xrightarrow{\uptimes \scalebox{1}{\textit{x}} \dtimes}$} $t$ $+\mathcal{L}_{\text{trans}}$ & 20.52 & \textbf{12.64} & 18.34 & 10.42 & 15.22 & 14.37 \\
        triangle $s$ \raisebox{-2pt}{$\xrightarrow{\uptimes \scalebox{1}{\textit{t}} \dtimes}$} $x$ & 20.38 & \textbf{12.40} & 18.50 & 10.22 & 15.62 & 14.77 \\
        triangle $s$ \raisebox{-2pt}{$\xrightarrow{\uptimes \scalebox{1}{\textit{t}} \dtimes}$} $x$ $+\mathcal{L}_{\text{trans}}$ & 20.64 & \textbf{12.42} & 19.20 & 10.21 & \textbf{15.87} & 14.89 \\
    \bottomrule
    \end{tabular}
    \caption{BLEU scores for each model and translation direction $s\rightarrow t$. In the multitask, cascade, and triangle models, $x$ stands for the third language, other than $s$ and $t$. In each column, the best results are highlighted. The non-highlighted results are statistically significantly worse than the single-task baseline.}
    \label{tab:bigtable}
\end{table*}
In all cases, the best models are the baseline single-task or simple multitask models. There are some instances, such as English-German, where the \textit{reconstruction} or the \textit{triangle} models are not statistically significantly different from the best model. The reason for this, we believe, is that in the case of text translation between so linguistically close languages, the lower level representations (the output of the encoder) provide as much information as the higher level ones, without the search errors that are introduced during inference.

A notable outcome of this experiment is that we do not observe the significant improvements with the \textit{reconstruction} models that \citet{tu2017neural} observed. A few possible differences between our experiment and theirs are: our models are BPE-based, theirs are word-based; we use Adam for optimization, they use Adadelta; our model has slightly fewer parameters than theirs; we test on less typologically different language pairs than English-Chinese.

However, we also observe that in most cases our proposed regularizers lead to increased performance. The \textit{invertibility} regularizer aids the \textit{reconstruction} models in achievινγ slightly higher BLEU scores in~3 out of the~6 cases. The \textit{transitivity} regularizer is even more effective: in~9 out the~12 source-target language combinations, the \emph{triangle} models achieve higher performance when trained using the regularizer. Some of them are statistical significant improvements, as in the case of French to English where English is the intermediate target language and German is the final target.

\section{Related Work}
The speech translation problem has been traditionally approached by using the output of an ASR system as input to a MT system. For example, \citet{ney1999speech} and \citet{matusov2005integration} use ASR output lattices as input to translation models, integrating speech recognition uncertainty into the translation model. 
Recent work has focused more on modelling speech translation without explicit access to transcriptions. \citet{long-EtAl:2016:NAACL-HLT} introduced a sequence-to-sequence model for speech translation without transcriptions but only evaluated on alignment, while \citet{anastasopoulos-chiang-duong:2016:EMNLP2016} presented an unsupervised alignment method for speech-to-translation alignment. 
\citet{bansal2017towards} used an unsupervised term discovery system \cite{jansen2010towards} to cluster recurring audio segments into pseudowords and translate speech using a bag-of-words model. 
\citet{berard2016listen} translated synthesized speech data using a model similar to the Listen Attend and Spell model \cite{chan2016listen}.
A larger-scale study~\cite{berard2018end} used an end-to-end neural system system for translating audio books between French and English.
On a different line of work, \citet{boito2017unwritten} used the attentions of a sequence-to-sequence model for word discovery.

Multitask learning \cite{caruana1998multitask} has found extensive use across several machine learning and NLP fields. For example, \citet{luong2015multi} and \citet{eriguchi2017learning} jointly learn to parse and translate; \citet{kim2017joint} combine CTC- and attention-based models using multitask models for speech transcription; \citet{dongetal2015multitask} use multitask learning for multiple language translation.
\citet{toshniwal2017multitask} apply multitask learning to neural speech recognition in a less traditional fashion: the lower-level outputs of the speech encoder are used for fine-grained auxiliary tasks such as predicting HMM states or phonemes, while the final output of the encoder is passed to a character-level decoder.

Our work is most similar to the work of \citet{weiss2017sequence}. They used sequence-to-sequence models to transcribe Spanish speech and translate it in English, by jointly training the two tasks in a multitask scenario where the decoders share the encoder. In contrast to our work, they use a large corpus for training the model on roughly~163 hours of data, using the Spanish Fisher and CALLHOME conversational speech corpora. The parameter number of their model is significantly larger than ours, as they use 8 encoder layers, and 4 layers for each decoder. This allows their model to adequately learn from such a large amount of data and deal well with speaker variation. However, training such a large model on endangered language datasets would be infeasible.

Our model also bears similarities to the architecture of the model proposed by \citet{tu2017neural}. They report significant gains in Chinese-English translation by adding an additional \textit{reconstruction} decoder that attends on the last states of the \textit{translation} decoder, mainly inspired by auto-encoders.

\section{Conclusion}

We presented a novel architecture for multitask learning that provides the second task with higher-level representations produced from the first task decoder. Our model outperforms both the single-task models as well as traditional multitask architectures. Evaluating on extremely low-resource settings, our model improves on both speech transcription and translation. By augmenting our models with regularizers that implement transitivity and invertibility, we obtain further improvements on all low-resource tasks.

These results will hopefully lead to new tools for endangered language documentation.
Projects like BULB aim to collect about $100$ hours of audio with translations, but it may be impractical to transcribe this much audio for many languages. For future work, we aim to extend these methods to settings where we don't necessarily have sentence triples, but where some audio is only transcribed and some audio is only translated.

\paragraph{Acknowledgements} This work was generously supported by NSF Award 1464553. We are grateful to the anonymous reviewers for their useful comments.

\bibliography{references}
\bibliographystyle{acl_natbib}

\end{document}